\documentclass[conference]{IEEEtran}
\IEEEoverridecommandlockouts
\usepackage{cite}
\usepackage{xurl}
\usepackage{amsmath,amssymb,amsfonts}
\usepackage{algorithmic}
\usepackage{graphicx}
\usepackage{textcomp}
\usepackage{xcolor}
\usepackage{subcaption}
\usepackage[hidelinks]{hyperref}
\usepackage{url}
\usepackage{acronym}

\def\BibTeX{{\rm B\kern-.05em{\sc i\kern-.025em b}\kern-.08em
    T\kern-.1667em\lower.7ex\hbox{E}\kern-.125emX}}
\begin{document}

\title{Multi-Label Perceptual Bug Detection in Video Games using Deep Learning on Gameplay Footage\\
}

\author{\IEEEauthorblockN{Nahian Rifaat}
\IEEEauthorblockA{
\textit{Ontario Tech University}\\
Oshawa, Ontario, Canada \\
nahian.rifaat@ontariotechu.net}
\and
\IEEEauthorblockN{Felix Morosov}
\IEEEauthorblockA{\textit{Universität Osnabrück } \\
Osnabrück, Niedersachsen, Germany \\
fmorosov@uni-osnabrueck.de}
\and
\IEEEauthorblockN{Loutfouz Zaman}
\IEEEauthorblockA{
\textit{Ontario Tech University}\\
Oshawa, Ontario, Canada \\
loutfouz.zaman@ontariotechu.ca}
}
\IEEEoverridecommandlockouts

\IEEEpubid{\makebox[\columnwidth]{979-8-3315-9476-3/26/\$31.00 ~\copyright2026 IEEE\hfill}
\hspace{\columnsep}\makebox[\columnwidth]{ }}
\maketitle
\IEEEpubidadjcol
\acrodef{FC}{Fully Connected}
\acrodef{AI}{Artificial Intelligence}
\acrodef{GELID}{GamEpLay Issue Detector}
\acrodefplural{GUI}[GUIs]{graphical user interfaces}
\acrodef{GUI}{graphical user interface}
\acrodef{DRL}{deep reinforcement learning}
\acrodef{RL}{reinforcement learning}
\acrodef{WOB}{World of Bugs}
\acrodef{ABD}{Automated Bug Detection}
\acrodef{JSON}{JavaScript Object Notation}
\acrodef{DBSCAN}{Density-Based Spatial Clustering of Applications with Noise}
\acrodef{LOD}{Level of Detail}
\acrodef{CFT}{Hybrid Co-Fine Tuning}
\acrodef{JEPA}{Joint Embedding Predictive Architecture}
\acrodef{ViT}{Vision Transformer}
\acrodef{fps}{frames per second}
\acrodef{ConvLSTM}{Convolutional Long Short-Term Memory}
\acrodef{RCNN}{Region-based Convolutional Neural Network}
\acrodef{BRNN}{Bidirectional Recurrent Neural Network}
\acrodef{LSTM}{Long Short-Term Memory Network}
\acrodef{ResNet-BiLSTM}{ResNet and Bidirectional Long Short-Term Memory}
\acrodef{BiLSTM}{Bidirectional LSTM}
\acrodef{ASL}{Asymmetric Loss}
\acrodef{I3D}{Inflated 3D ConvNet}
\acrodef{R3D-18}{3D ResNet}
\acrodef{GRU}{Gated Recurrent Unit}
\acrodef{DLC}{Downloadable Content}

\begin{abstract}
Traditional approaches for automated bug detection in video games, such as manual testing, can be beneficial for the improvement of quality assurance, but they can be expensive and time-consuming. The scarce number of tools available to detect multiple perceptual bugs in the same video frame introduces detection challenges for automated bug detection tools in real-world scenarios. We propose a deep learning model for multi-label perceptual bug detection and compare it against video classification models such as Inflated 3D ConvNet and 3D ResNet. Our proposed model, ResNet-BiLSTM, achieved an F1 score of 85.78\% on the benchmark dataset. Our results demonstrated that temporal dependency modelling is beneficial for accurate video-based bug detection. We believe this work with multi-label perceptual bug detection on gameplay videos will help save resources spent on manual testing workloads in video games. Furthermore, we introduce a new dataset with multi-label perceptual bugs in this work. The dataset contains 77,969 video clips across different genres of games with approximately 1.2 million frames, containing combinations from 5 classes of bugs in the same video frame. 
\end{abstract}

\begin{IEEEkeywords}
Automated bug detection, bidirectional long short-term memory (Bi-LSTM), multi-label classification.
\end{IEEEkeywords}

\section{Introduction}
While playing video games, encountering a bug can reduce the quality of immersion a player experiences \cite{ariyurek2019automated}. There are numerous effects of these bugs causing minor to major negative impacts on player experience, the reputation of the developers of the game, and the loss of potential revenue that can prevent players from reaching their intended goal \cite{Rollings2003}. 

Although games are thoroughly tested with massive amounts of resources directed towards quality assurance before launch \cite{albaghajati2020video, gutierrez2021proposal}, even AAA game developers face the reality of bugs creeping into their games \cite{politowski}. The Last of Us Part 1 faced such issues with character models missing details, jittery camera movement, and distorted hair, among others \cite{Sinha2023}. These bugs damaged the reputation of the developers and the publishing company. The scale and intricacies of such games present a real challenge to ensure players have a bug-free experience before release. Furthermore, manual testing and other effective methods of detecting bugs can be time-consuming and expensive \cite{albaghajati2020video}. Coppola et al. \cite{knowyourbugs} proposed a taxonomy of 26 distinct issues and combined them into five macro-categories. They reported that available research focuses a lot on bugs detectable through the multimedia appearance of a game, such as animation errors, HUD failures, and glitches in rendering. The authors noted that automated detection requires complex and resource-heavy instrumentation of the game engine. This underscores the need for detection methods that can identify bugs without requiring deep integration into the game engine. The vastness of the variety of bugs can make it almost impossible and impractical to search for bugs at every level for complex and expansive video games \cite{Roque2025}. 

Among low-level, engine, application, perceptual, and behavioural bugs, the latter two cannot be detected with logical rules \cite{astrobug}. Perceptual bugs in games are issues that affect the display or appearance of game elements \cite{astrobug}. On the contrary, behavioural bugs have a negative impact on game mechanics or game logic. In the era of \ac{AI}, game developers can use deep learning to automate and effectively enhance the bug detection process \cite{rl, playtest-deep}. 

Perceptual bugs present unique challenges for bug detection. These classes of bugs, such as geometry corruption, texture corruption, and  Z-clipping, require an understanding of visual characteristics. Furthermore, perceptual bugs such as Z-fighting bugs cause objects to flicker, which requires an understanding of the bug's temporal characteristics in addition to its visual ones over multiple frames \cite{wilkins2022learning}. Additionally, video clips might contain multiple bugs occurring at the same time due to cascading defects, introducing the challenge of identifying multiple simultaneous perceptual bugs, which is more difficult than multiclass classification.

To address the mentioned difficulties, we framed perceptual bug detection as a supervised multi-label classification problem. In this work, we introduce a new architecture combining spatial and temporal feature extraction to create an automated test oracle to solve the multi-label bug classification problem. Our key contributions in this work are as follows:

\begin{itemize}
    \item a new model architecture combining spatial and temporal feature extraction that learns a multi-task objective for multi-label classification;
    \item a benchmark dataset \cite{rifaat2026perceptualbugdataset} for multi-label perceptual bug detection with 77,969 video clips prepared from recorded gameplay sessions, containing clips with a combination of multiple perceptual bugs (up to a combination of 3 classes of bugs from a set of 5 classes of perceptual bugs).
\end{itemize}

\section{Related Work}
\subsection{Traditional Game Testing}
In traditional game testing, human testers participate in sessions where they conduct manual assessments and observations to recognize bugs. Human testers actively look for potential bugs while they execute test cases. Usability testing evaluations are widely used in these manual approaches. 

For example, a case study was conducted \cite{manualtest1} where Choi was able to identify that usability expert evaluation provides valuable and novel feedback for testing games. Korhonen and Koivisto \cite{manualtest2} proposed a model including three main modules, namely, usability, mobility, and gameplay, for playability heuristics for mobile games. In this study, the model's effectiveness in enhancing the game evaluation and playability issue detection was tested and verified. Moreno-Ger et al. \cite{manualtest3} proposed a new methodology based on analysis of recorded gameplay that provided a systematic and targeted approach to suggest improvements that benefited playtesting in serious games. Diah et al. \cite{manualtest4} conducted a study to verify the effectiveness of a game called Jelajah within their proposed usability testing method with children. Their data proved that their proposed evaluation method can be effectively adopted for usability testing. Duarte et al. \cite{11028401} introduced a systematic framework for exploratory testing. The framework consisted of nine tailored strategies. This framework successfully helped uncover 80 bugs in the study. However, the authors confirmed that such methods are heavily reliant on human testers, which makes evaluation costly and time-consuming. Additionally, all these manual testing strategies could be used to detect perceptual bugs, but would lead to the same outcome in resource expenses.

\subsection{Automated Bug Detection}
\ac{ABD} primarily involves software agent creation with the ability of game engagement and exploration to uncover and identify bugs. Although there is no definitive answer to the superior effectiveness of \ac{ABD} techniques compared to traditional methods such as manual testing, earlier researchers have conducted comparative studies in this regard. Dobles et al. \cite{abd1} compared manual and automated testing and found that automated tests require more initial effort compared to manual ones. However, their findings suggest that automated testing is superior for repetitive testing and performs better in defect detection.

Varvaressos et al. \cite{runtime} reported an approach where instrumenting the game loop could enable game developers to monitor the temporal properties in the runtime and detection violations of formal specifications during gameplay. Such methods require access to the game's state and manual definition of rules.


Iftikhar et al. \cite{abd2} introduced an automated functional testing approach using a model-based methodology for games and evaluated it on two platformer games. Their proposed approach employed unified modelling language profiles, class diagrams, and state machines to automatically generate and execute test cases. The authors reported that the effectiveness of their approach might be less effective in complex games because of its inability to cover all paths and functionalities. This approach requires users with previous software engineering experience to create models prior to testing.

Lovreto et al. \cite{abd3} manually crafted some functional test scripts for their proposed automated test script method and evaluated the method on 16 mobile games. However, the authors reported that their approach had several limitations, including the inability to address the unpredictability inherent in certain aspects of the games effectively.

Guglielmi et al. \cite{abd4} proposed \ac{GELID} for anomaly identification through the analysis of gameplay videos to offer insights to game developers to improve their games. To evaluate \ac{GELID}, they performed a study on 604 video segments derived from 80 hours of gameplay footage across three video games. The authors reported that the segmentation process and issue-based clustering were effective. However, the steps involving categorization and context-based clustering segments are reported to require refinement before being implemented in practical applications reliably.

Prasetya et al. \cite{abd5} proposed an agent-based automated navigation framework for playtesting in video games. The agents perform graph-based pathfinding. The framework uses search-based and rule-based techniques. However, the authors reported that the framework lacks generalization.

Guerrero-Romero et al.~\cite{romero} proposed an automated testing approach using a team of AI agents to detect behavioural anomalies and validate game design during development.

Albaghajati and Ahmed \cite{abd6} leveraged two agent-based genetic algorithms for playtesting on video games. One of the agents generates buggy states within the game, while the other agent analyzes the game state using a representation based on Petri nets modelling. The approach relies on rule-based logic and struggles to identify a bug from which other related bugs emerge. The authors acknowledge that their algorithm struggles to effectively pinpoint the primary bug that could potentially create additional issues in the game.

Zhao et al. \cite{abd8} introduced an approach that derived playing tactics from actual gameplay and uses them to automatically test games for potential bugs. This approach incorporates a library of rule-based inferences and is able to respond to randomly generated game scenarios. But the applicability of this framework may be limited in more complex games.  
\subsection{AI and Machine Learning in Game Testing}
\ac{AI} and Machine Learning (ML) has proven to be a powerful tool for bug detection backed by years of previous literature. With modern video games increasing in challenges and growing complexity, and a requirement for quicker release cycles and \ac{DLC} updates, manual playtesting, along with other traditional methods of bug detection, continues to remain resource-intensive and expensive. Approaches that incorporate \ac{AI} are a promising alternative for \ac{ABD} with more efficiency and effective results.

Bug identification and automated game exploration are the two main components of \ac{ABD}. Wuji is one of the most recent frameworks that incorporates \ac{ABD} and game exploration with the use of \ac{DRL} \cite{wuji}. Wuji detects four categories of bugs, namely, crash, stuck, logical, and gaming balance, by combining evolutionary multi-object optimization and DRL. Rani et al. \cite{Rani2022} demonstrated that \ac{DRL} can be used to find the presence of bugs by analyzing game screens effectively even in low-quality or blurred frames. Xue used Q-learning to explore and test \acp{GUI} for a collection of mobile apps to create behaviour models and train two optimization objectives, namely, activity coverage and the number of crashes \cite{sigsoft}. However, the methods proposed by Zheng et al. \cite{wuji} and Xue \cite{sigsoft} were both rule-based approaches that could not detect high-level perceptual and behavioural bugs. In this work, our focus is to design models capable of detecting these challenging bugs faced by players. Furthermore, a self-learning game testing framework that can navigate and take advantage of game mechanics through \ac{RL} guided by a predefined reward signal was introduced by Bergdahl et al. \cite{rl}. Pfau et al. \cite{chiplayicarus} focused on developing an autonomous framework, where games are played, and bugs are reported. Pfau et al. used discrete \ac{RL}, where they used short-term and long-term memory in pairs as a solving mechanic. The mentioned literature mostly works on game environment exploration rather than bug detection. On the contrary, our work detects bugs on a benchmark dataset created from human-played sessions and works towards maximizing detecting multiple bugs at the same time in gameplay videos.

Wilkins et al. \cite{world} introduced an open platform for ABD testing for 3D game environments, focusing more on perceptual bugs. The platform in this study called \ac{WOB}, employs learning-based methods by leveraging the rendered scenes. An \ac{AI} agent explores, captures, and detects bugs from the captured video frames from the game environment, similar to a real player. The authors identified ten types of perceptual bugs and varying detection accuracy across categories, probably stemming from instances in which the object's rear geometry did not render properly, tricking the model into not classifying the event as a bug. 

Paduraru et al. \cite{anomaly} introduced a hierarchical detection framework using real-time game state information to identify visual anomalies, addressing a range of glitches, namely, mesh distortions, missing textures, and camera clipping. In this work, they used a two-stage hierarchical architecture. The first stage used a ShuffleNetV2 model to identify regions of interest, and in the second stage, another backbone is triggered for verification of the bug if an anomaly is suspected. Their system is designed to stream screenshots to external inference servers to avoid impacting the game's runtime performance. Earlier, we framed bug detection as an anomaly detection problem, making the system learn the normal state of a game and identify deviations as bugs  \cite{astrobug, astromini}. We integrated an \ac{RL} agent to autonomously explore game environments and gather data. We used a \ac{ConvLSTM} model to process sequences of gameplay frames and captured temporal dependencies by a sliding window approach that helped detect bugs such as flickering or inconsistent rendering. The framework is trained on bug-free gameplay, and once an anomaly is detected, the framework employs the bug classification stage and uses the  \ac{DBSCAN} algorithm to cluster predicted buggy frames into distinct categories of perceptual bugs, such as geometry corruption, boundary hole, Z-clipping, and so on. In summary, our previous work first detected anomalies and then used unsupervised clustering to identify the type of perceptual bug. Lin et al. \cite{lin2019identifying} introduced a methodology for video-based bug detection where they used video metadata and the random forest classifier to rank gameplay videos based on the likelihood of bug presence by framing bug detection as a binary classification problem. 

Ling et al. identified rendering defects such as stretched geometry and missing textures with a ShuffleNetV2 architecture on static individual frames \cite{ling2020using}, but only dealt with a single visual glitch each frame. Abdelfattah et al. detected multi-frame bugs such as \ac{LOD} pops and culling pops in their work using a \ac{CFT} framework to detect visual bugs when there is a scarcity of labelled data \cite{cofinetuning}. They defined the task as an object detection objective to both localize using bounding boxes and classify visual bugs within a frame.  The model was trained simultaneously on labelled data from the "target" game and a mixture of other games. To utilize large amounts of unlabelled data, they used a \ac{JEPA}, which involves masking image patches and tasking the model with reconstructing them in the embedding space. Their framework utilizes a faster \ac{RCNN} object detection architecture with a \ac{ViT} backbone. Savran et al. \cite{yolo} used YOLO v10 to detect visual and logical bugs to perform real-time localization of bugs, treating bug detection as an object detection problem. The dataset used in this work has images for each bug class out of five classes of bugs. 

\begin{table}[htbp]
\centering
\scriptsize
\caption{Comparison of automated bug detection approaches}
\label{table:comparison}
\resizebox{\columnwidth}{!}{%
\begin{tabular}{l l c l}
\hline\hline
\textbf{Study} & \textbf{Approach} & \textbf{Multi-label} & \textbf{Data} \\
\hline
Varvaressos et al. \cite{runtime} & Runtime Monitoring & No & Engine Hooks \\
Lin et al. \cite{lin2019identifying} & Ranking & No & Video metadata \\
Abdelfattah et al. \cite{cofinetuning} & Hybrid co-finetuning & No & Multiframe \\
Rani et al. \cite{Rani2022} & DRL (DQN) & No & Image \\
Savran et al. \cite{yolo} & YOLO v10 & No & Image \\
Paduraru et al. \cite{anomaly} & ShuffleNetV2 & No & Image+State \\
Our previous study \cite{astrobug, astromini} & Anomaly detection & No & Video \\
\textbf{This study} & \textbf{ResNet-BiLSTM} & \textbf{Yes} & \textbf{Video} \\
\hline\hline
\end{tabular}%
}
\end{table}

As illustrated in Table~\ref{table:comparison}, each literature approached bug detection in games differently. However, none of the mentioned literature tackled the problem of detecting multiple perceptual bugs at the same time in the gameplay footage. To the best of our knowledge, such a dataset is not available. Therefore, we acknowledged this research gap and developed a benchmark dataset with large amounts of labelled data, which consists of human-played gameplay footage with multiple perceptual bugs occurring simultaneously. Furthermore, we developed a new model architecture to test our proposed model's performance on the mentioned dataset.

\section{Materials and Methods}
This section covers the details of the benchmark datasets used in this study. The dataset contains data collected from gameplay footage. We have also provided the details of the proposed neural network architecture used in this study for multi-label perceptual bug detection on the mentioned benchmark dataset.

\begin{figure}[h!]
    \centering
    \includegraphics[width=\columnwidth, height=0.6\textheight, keepaspectratio]{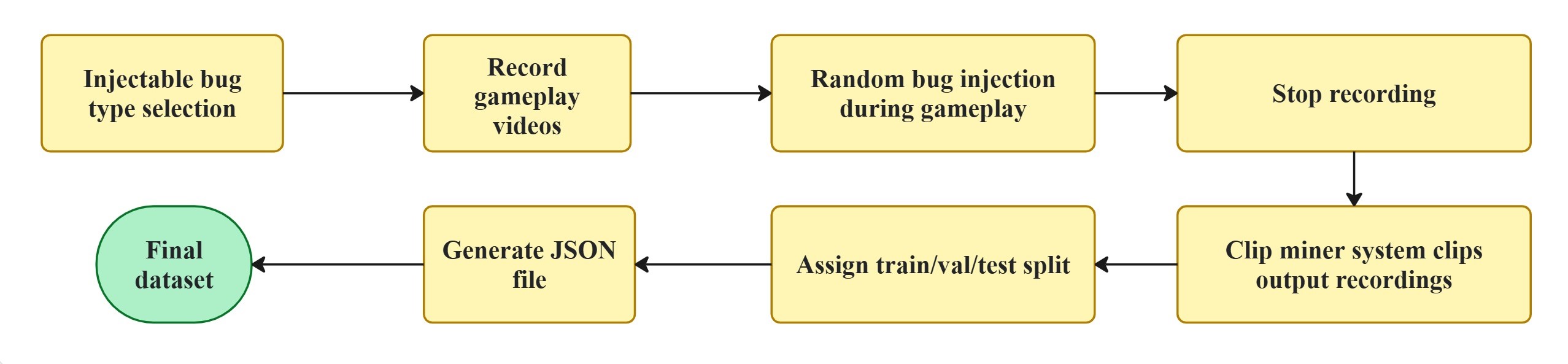}
    \caption{Dataset preparation pipeline.}
    \label{fig:dataset-preparation}
\end{figure}

\subsection{Dataset Preparation}
The benchmark dataset was prepared by recording gameplay sessions from three different open-source games, namely, Open Nights \cite{open}, Unity3D Mario Kart Racing Game \cite{mkart}, and FPS Microgame \cite{fps}. The data collection procedure for the study was approved by the institutional Research Ethics Board (Reference File No.: 20215). The games selected for data collection are described below:
 
\subsubsection{Open Nights}
Open Nights is a first-person horror game focused on survival. This game is characterized by low-light environments and static camera angles. Open Nights was chosen for the dataset to test the model's ability to detect bugs that are harder to detect in low light, such as texture corruption and Z-fighting.
\subsubsection{Unity3D Mario Kart Racing Game}
The Unity3D Mario Kart Racing Game is an open-source implementation of the Mario Kart game. It is a third-person racing game with high-speed movements on colourful racing tracks. We selected this game to test models on bugs in complex 3D environments, where bugs occur with high-speed moving assets.
\subsubsection{FPS Microgame}
The FPS Microgame is a fast-paced first-person shooter game characterized by rapid camera movements. This game was selected for the dataset to test the ability of models to distinguish motion blur from bugs such as geometry corruption and Z-fighting.

The above titles were selected so that our dataset contained footage from diverse genres of games with different graphical demands and bugs to test our models on.

\begin{figure}[ht!]
    \centering
    \begin{subfigure}[b]{0.49\columnwidth}
        \includegraphics[width=\linewidth]{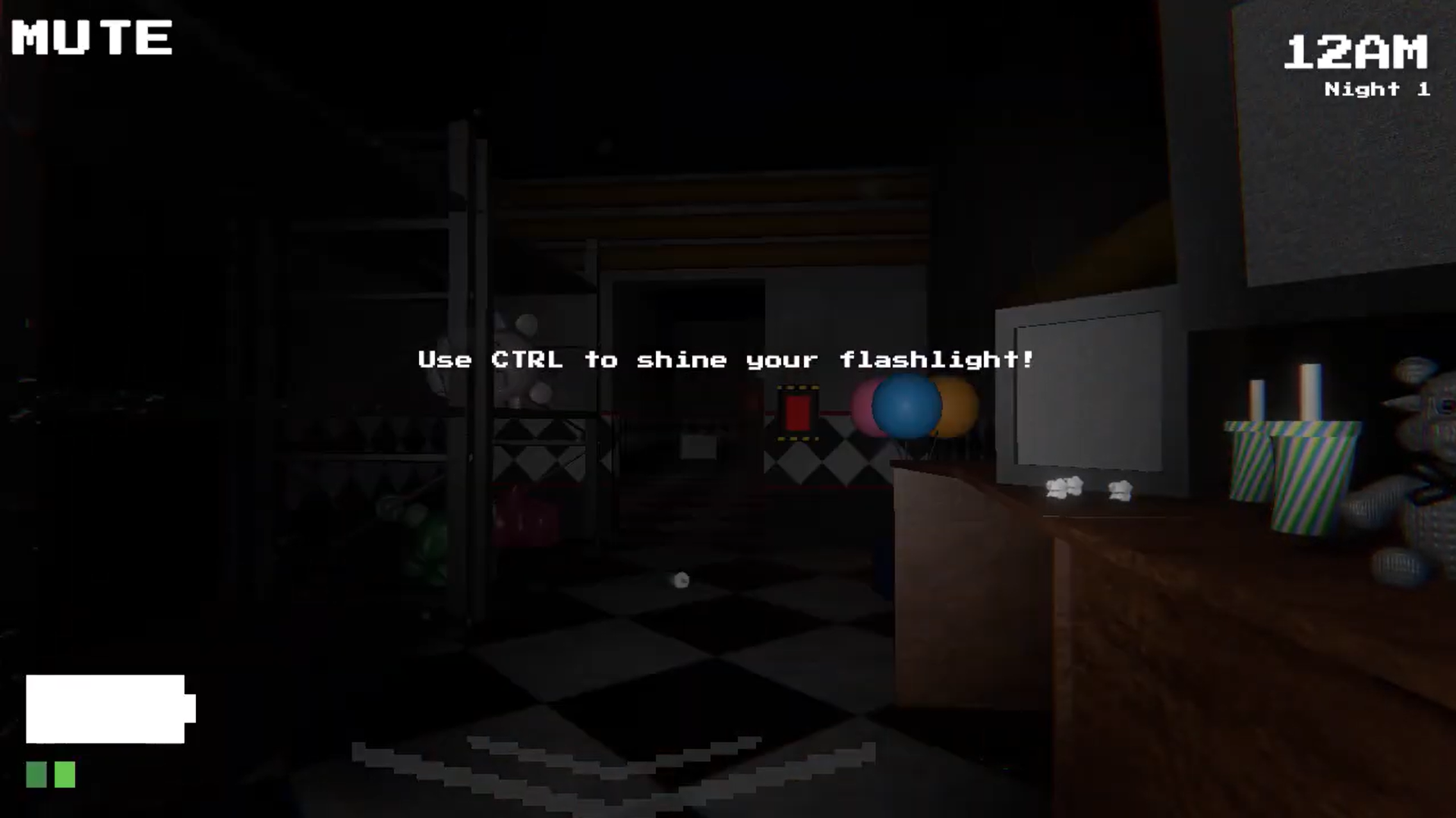}
        \caption{Z-clipping in the Open Nights game: before}
        \label{fig:before-zclip}
    \end{subfigure}
    \hfill
    \begin{subfigure}[b]{0.49\columnwidth}
        \includegraphics[width=\linewidth]{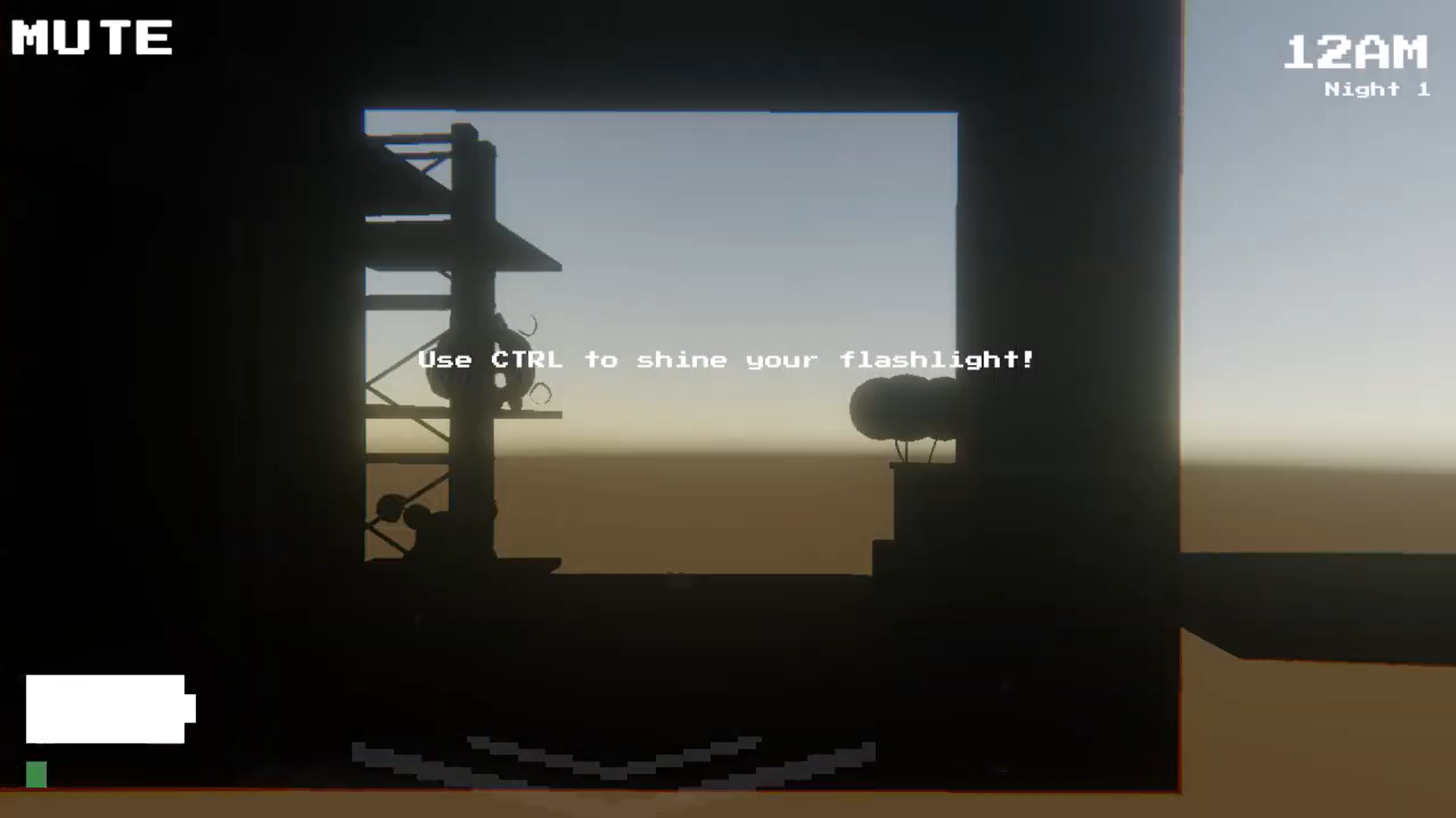}
        \caption{Z-clipping in the Open Nights game: after}
        \label{fig:zclip}
    \end{subfigure}

 \vspace{0.5em}

    \begin{subfigure}[b]{0.49\columnwidth}
        \includegraphics[width=\linewidth]{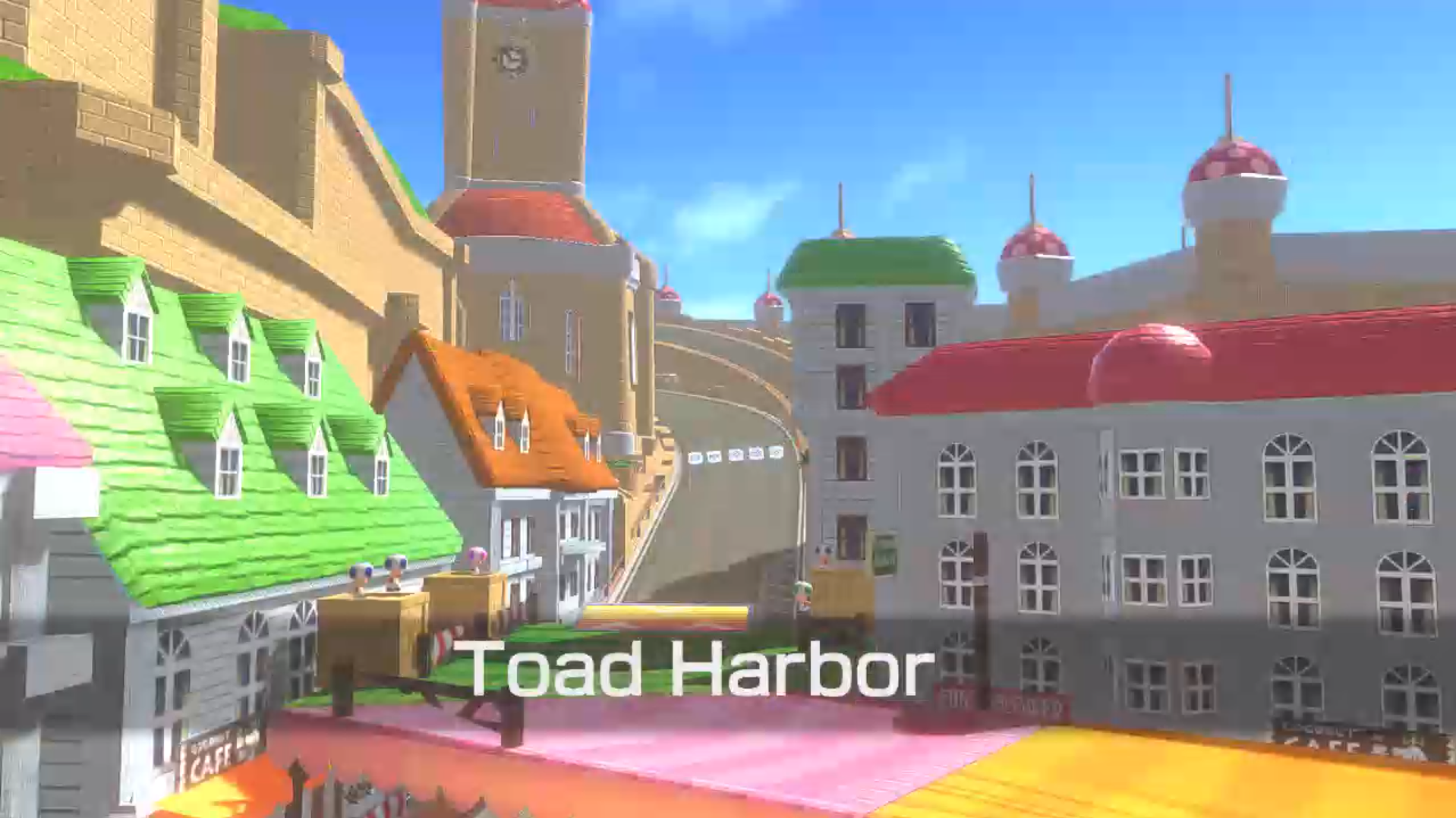}
        \caption{Texture corruption in Mario Kart: before.}
        \label{fig:before-texture}
    \end{subfigure}
    \hfill
    \begin{subfigure}[b]{0.49\columnwidth}
        \includegraphics[width=\linewidth]{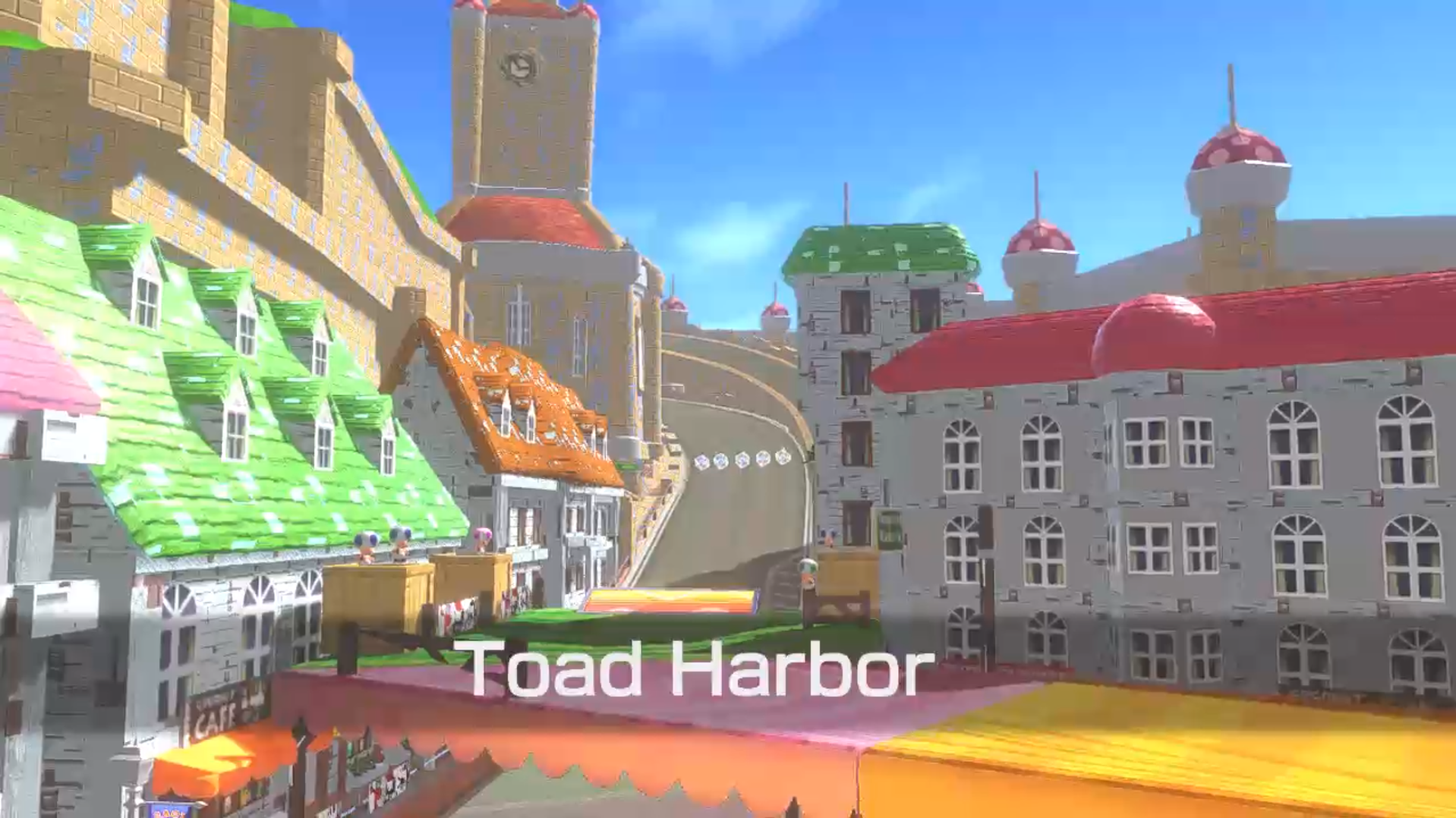}
        \caption{Texture corruption in Mario Kart: after.}
        \label{fig:corrupted-texture}
    \end{subfigure}

    \vspace{0.5em}

    \begin{subfigure}[b]{0.49\columnwidth}
        \includegraphics[width=\linewidth]{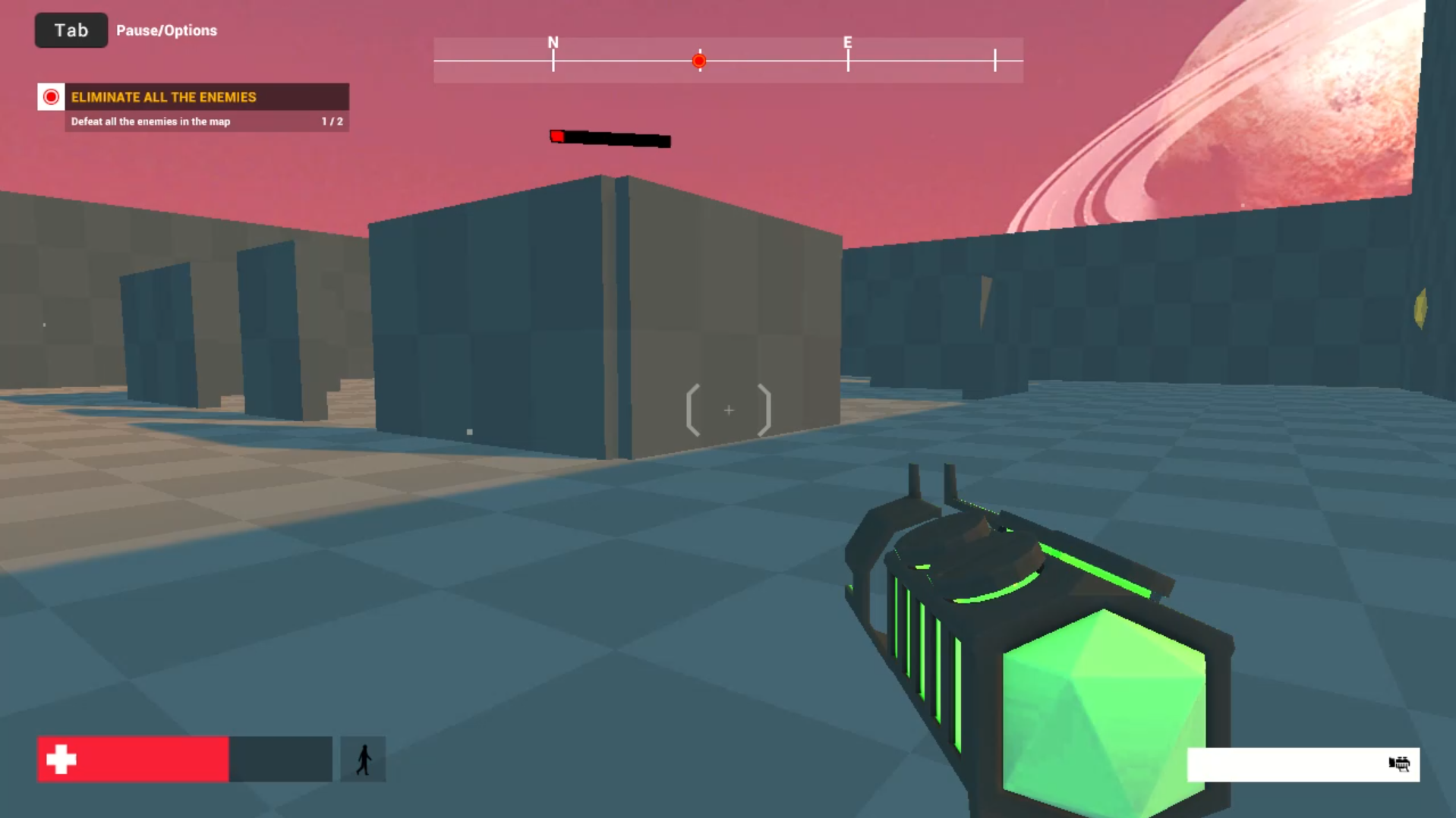}
        \caption{Geometry corruption in the FPS Microgame: before.}
        \label{fig:before-geocorruption}
    \end{subfigure}
    \hfill
    \begin{subfigure}[b]{0.49\columnwidth}
        \includegraphics[width=\linewidth]{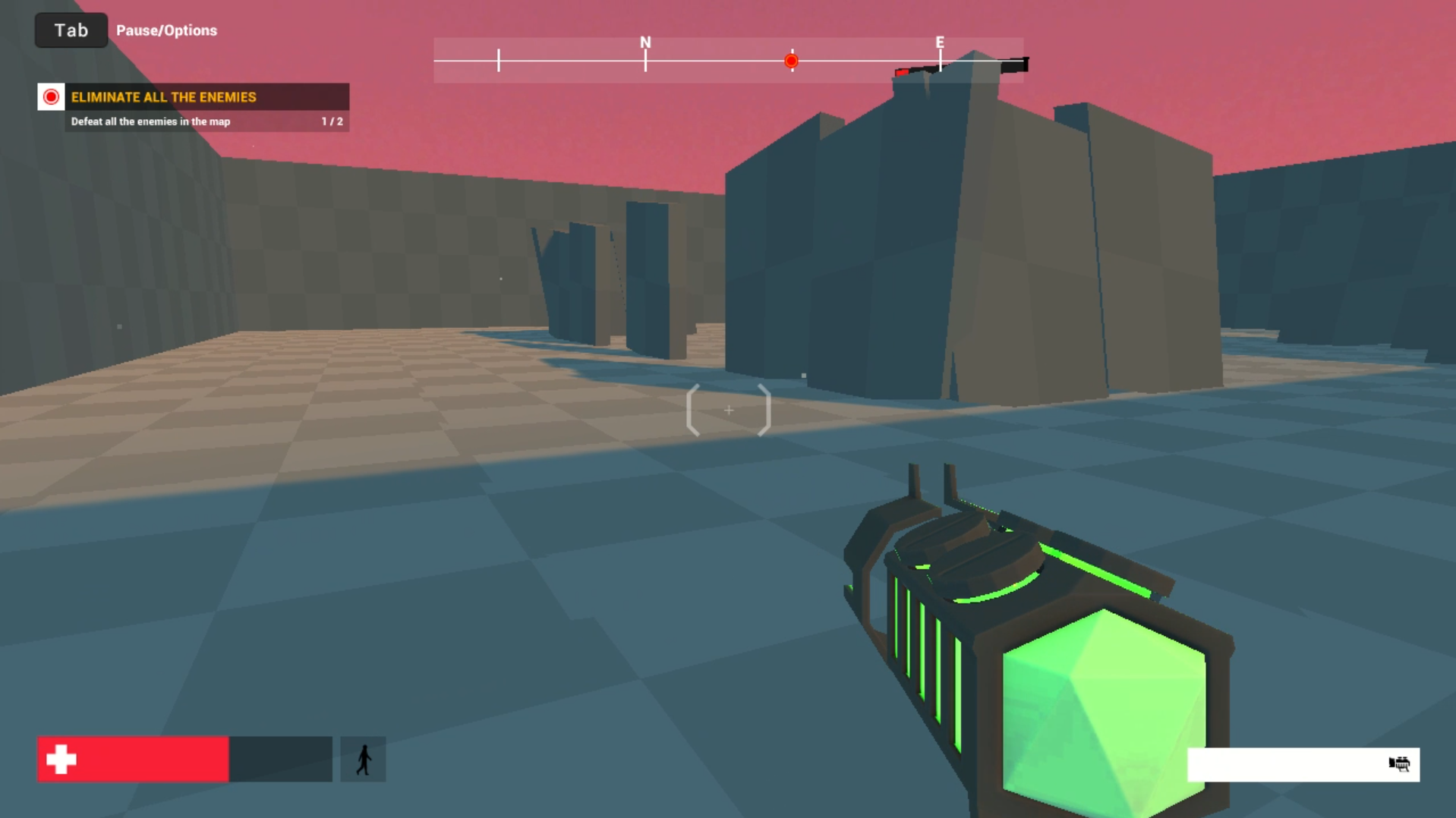}
        \caption{Geometry corruption in the FPS Microgame: after.}
        \label{fig:geocorruption}
    \end{subfigure}

    \caption{Examples of perceptual bugs from the dataset.}
    \label{fig:bug-examples}
\end{figure}

11 participants volunteered to be part of the gameplay recording sessions and were recruited through convenience sampling. The videos were recorded at 30 \ac{fps}. To introduce bugs, we deployed a bug injector system and a gameplay recorder within the Unity projects of the mentioned games to record the gameplay sessions of the participants. During the gameplay sessions, the bug injector algorithm introduced perceptual bugs at random moments or when the game character came in close range of a predetermined game object. We chose to work with artificially introduced bugs because the chosen games were open-source and did not have perceptual bugs. Up to 3 perceptual bugs were injected at the same time from a list of 5 classes of perceptual bugs, namely, geometry corruption, texture corruption, boundary hole, Z-fighting, and Z-clipping. These bug types have distinct features, which allows the creation of models that can classify them properly. Furthermore, all of the classes of bugs are injected into the gameplay sessions of each game. After the recording of a gameplay session was finished, our clip miner system resized the gameplay footage to a resolution of $224 \times 224$ and separated the recorded video into 2-second clips at 8 \ac{fps}. After all of the gameplay sessions had been recorded, we froze the dataset by splitting it into train (70\%), validation (15 \%), and test (15 \%) sets, maintaining the same proportions of the number of bug combinations (up to three bugs) in each split. The dataset was finalized by preparing a \ac{JSON} file that contained the data of the clips, the dataset splits, and the data about the bugs within those video clips. The dataset preparation pipeline is shown in Fig.~\ref{fig:dataset-preparation}.
\begin{figure*}[h!]
    \centering
    \includegraphics[width=\textwidth, height=0.5\textheight, keepaspectratio]{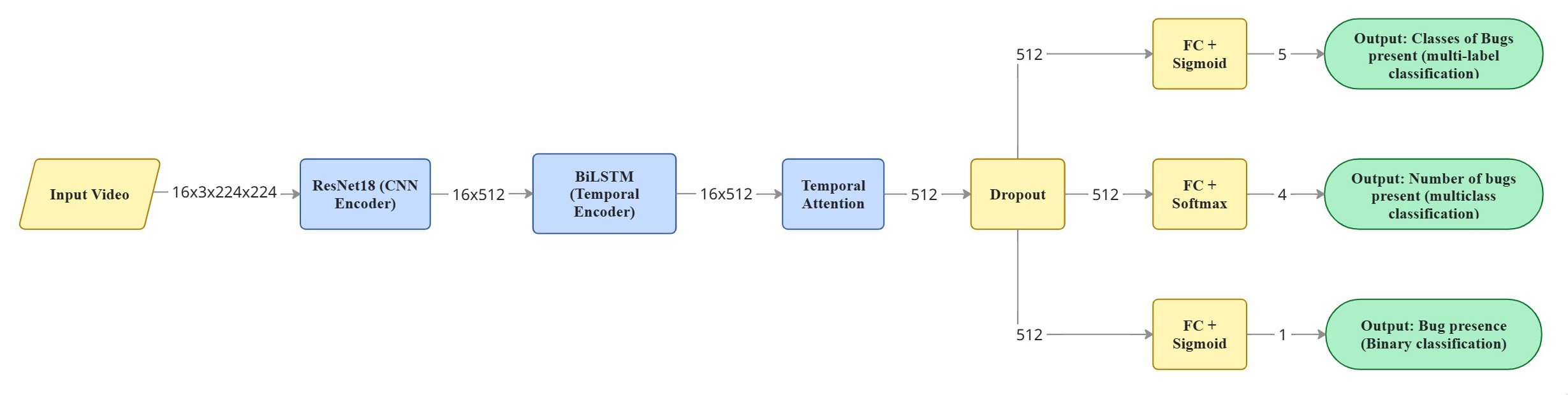}
    \caption{ResNet-BiLSTM architecture.}
    \label{fig:resnet-bilstm}
\end{figure*}

\subsection{Dataset Description}\label{sub:dataset-description}
The multi-label bug detection dataset we prepared consisted of five classes of perceptual bugs (bugs that affect the appearance of game elements), up to a combination of three concurrent bugs per clip taken from the gameplay footage. We set the bug combinations to a maximum of three concurrent perceptual bug classes per frame to prevent overlapping bugs from occluding each other, allowing models to accurately recognize the visual features of each class of bug. Some examples of the bugs available in the dataset are shown in Fig.~\ref{fig:bug-examples}. The five classes of bugs are, namely, boundary hole, texture corruption, geometry corruption, Z-clipping, and Z-fighting. Each bug type occurs at most once in a video clip. Gameplay sessions from all 3 games are injected with each class of the mentioned bugs.

The dataset consists of 77,969 video clips (2-second clips at 8 \ac{fps}) prepared from the gameplay sessions. The dataset has a total of 1,247,504 frames and includes all the video clips. This dataset was split into train (70\%), validation (15\%), and test (15\%) sets. Each split contains all 5 classes of perceptual bugs previously mentioned and approximately the same proportion of the number of bug combinations (up to three bugs per video clip). Further details about the dataset are shown in Table~\ref{tab:bug-count-combined} and Table~\ref{tab:bug-type-distribution}.

\begin{table}[h!]
    \centering
    \caption{Distribution of bug count combinations across splits.}
    \label{tab:bug-count-combined}
    \begin{tabular}{lccc}
        \hline\hline
        \textbf{Bug Count} & \textbf{Test} & \textbf{Train} & \textbf{Validation} \\
        \hline
        0 bug(s) & 7,669 (65.55\%) & 35,785 (65.57\%) & 7,668 (65.58\%) \\
        1 bug(s) & 2,103 (17.97\%) &  9,807 (17.97\%) & 2,101 (17.97\%) \\
        2 bug(s) & 1,070 (9.15\%)  &  4,986 (9.14\%)  & 1,068 (9.13\%)  \\
        3 bug(s) &    858 (7.33\%) &  3,998 (7.33\%)  &    856 (7.32\%)  \\
        \hline
        \textbf{Total} & \textbf{11,700} & \textbf{54,576} & \textbf{11,693} \\
        \hline\hline
    \end{tabular}
\end{table}

\begin{table}[ht!]
    \centering
    \scriptsize
    \caption{Individual bug class distribution in videos across splits.}
    \label{tab:bug-type-distribution}
    \begin{tabular}{lcccc}
        \hline\hline
        \textbf{Bug Type} & \textbf{Test} & \textbf{Train} & \textbf{Validation} \\
        \hline
        Boundary Hole       & 1{,}313 (11.22\%) & 6{,}138 (11.25\%) & 1{,}363 (11.66\%) \\
        Texture Corruption   & 1{,}434 (12.26\%) & 6{,}616 (12.12\%) & 1{,}352 (11.56\%) \\
        Geometry Corruption & 1{,}287 (11.00\%) & 6{,}205 (11.37\%) & 1{,}326 (11.34\%) \\
        Z-Clipping          & 1{,}450 (12.39\%) & 6{,}620 (12.13\%) & 1{,}413 (12.08\%) \\
        Z-Fighting          & 1{,}333 (11.39\%) & 6{,}194 (11.35\%) & 1{,}351 (11.55\%) \\
        \hline\hline
    \end{tabular}
\end{table}

\subsection{Model Architecture}
In this study, we introduce a model architecture, \ac{ResNet-BiLSTM}, for the multi-label bug detection task. The model first extracts visual features from the frames of the video using a pre-trained ResNet-18 \cite{resnet} and then feeds the extracted features into a \ac{BRNN} \cite{Schuster1997}. In our case, we are using a \ac{LSTM} \cite{lstm} to use within the  \ac{BRNN}. This \ac{BiLSTM} network is used to extract temporal information from the video frames. Next, the output of the \ac{BiLSTM} is fed into the Temporal Attention layer to extract the most essential features for the multi-label classification task, followed by a dropout regularization layer. The output of the dropout layer is then fed into three separate classification heads. The first classification head has a \ac{FC} layer followed by a sigmoid activation function to classify the multiple bugs present in the video clip. Furthermore, the second classification head consists of an \ac{FC} layer followed by a softmax classifier to predict if the video contains the number of bugs between 0, 1, 2, or 3 bugs. Finally, the third classification head contains an \ac{FC} layer followed by a sigmoid classifier to detect the presence of bugs. The model architecture with output shapes after each layer is shown in Fig.~\ref{fig:resnet-bilstm}.

\section{Experimental Setup}
This section covers the setup used in the experiments, including the environment, hyperparameters and evaluation metrics.
\subsection{Hardware, Environment and Hyperparameters}
The experiments ran on servers with an NVIDIA H100 GPU. The server ran on the Fedora Linux operating system. We used PyTorch to build the models and experiments. Furthermore, we used Weights \& Biases to track the experiments. Training hyperparameters for our model are given below:

\begin{itemize}
    \item \textbf{Epochs:} 60
    \item \textbf{Batch size:} 32
    \item \textbf{Dropout rate:} 0.3
    \item \textbf{Learning rate:} 0.0007
    \item \textbf{Optimizer:} ADAM \cite{kingma2014adam}
\end{itemize}

\begin{figure*}[h!]
    \centering
    \includegraphics[width=\textwidth, height=0.9\textheight, keepaspectratio]{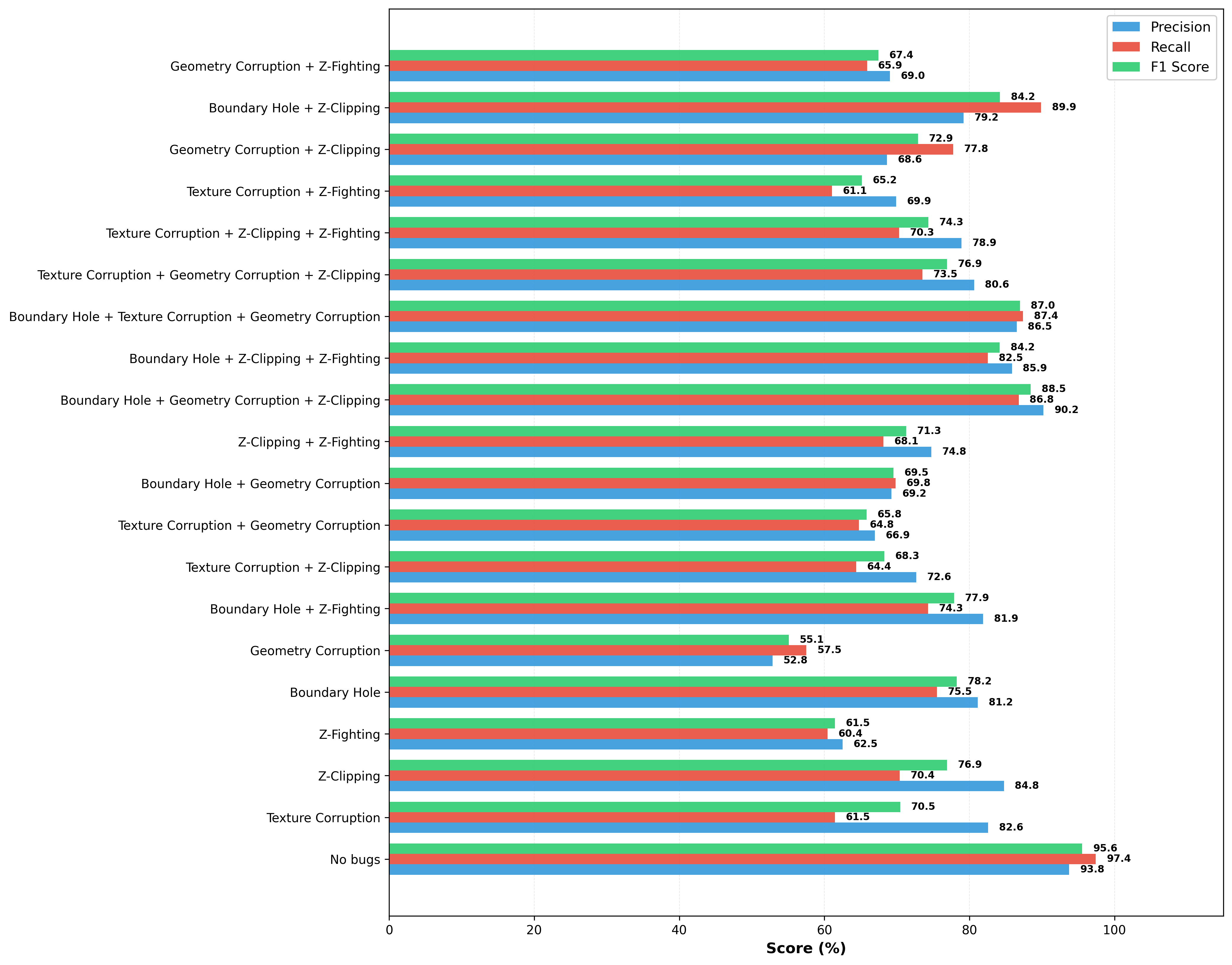}
    \caption{Evaluation metrics per bug combination for the proposed ResNet-BiLSTM model on the test set.}
    \label{fig:bug-combination-metrics}
\end{figure*}

\subsection{Loss function}
For the model architecture shown in Fig.~\ref{fig:resnet-bilstm}, we used a combination of different loss functions. The multi-task objective used for our multi-label bug classification task, which includes a combination of three loss terms, is as follows:

\begin{equation}
    \mathcal{L} = \mathcal{L}_{\text{types}} + \lambda_{\text{cnt}}\,\mathcal{L}_{\text{count}} + \lambda_{\text{any}}\,\mathcal{L}_{\text{any}} 
    \label{eq:total-loss}
\end{equation}
where $\lambda_{\text{cnt}}$, $\lambda_{\text{any}}$
are scalar weights controlling the contribution of each auxiliary term. \noindent$\mathcal{L}_{\text{types}}$ is the \ac{ASL} ~\cite{ridnik2021asymmetric} 
being used for multi-label bug classification. \noindent$\mathcal{L}_{\text{count}}$ is the Focal Cross-Entropy~\cite{lin2017focal} loss 
being used for bug count detection, and \noindent$\mathcal{L}_{\text{any}}$ indicates the Binary Cross-Entropy loss for bug presence detection. 

\subsection{Evaluation metrics}
We use the micro-averaged F1 score as the evaluation metric 
for multi-label bug type classification because micro-averaging gives equal weight to each prediction rather than each class, making it suitable for imbalance among classes in the dataset and also because there is a dominance of examples with no bugs. The equation for the micro-averaged F1 score is as follows:
\begin{equation}
    F_1 = \frac{2 \times P \times R}{P + R}
    \label{eq:f1}
\end{equation}

where Precision, $P = \frac{TP}{TP + FP}$ and Recall $R = \frac{TP}{TP + FN}$, with $TP$, $FP$, and $FN$ denoting true positives, false positives, and false negatives, respectively.

\section{Results}
This section summarizes the outcomes of the experiments performed utilizing the proposed architecture from our study on the proposed dataset. We compared the proposed model with video-based classification models.

\begin{table}[h!]
    \centering
    \caption{Comparison of different model architectures on the test dataset.}
    \label{tab:model-comparison}
    \begin{tabular}{lc}
        \hline
        \textbf{Model} & \textbf{F1 Score (\%)} \\
        \hline
        I3D \cite{carreira2017quo}                         & 19.54 \\
        R3D-18 \cite{tran2018closer}                     & 18.01 \\
        ResNet-18 + GRU              & 19.34 \\
        \textbf{ResNet-18 + BiLSTM (Ours)} & \textbf{85.78} \\
        \hline
    \end{tabular}
\end{table}

As shown in Table~\ref{tab:model-comparison}, the proposed model architecture shown in Fig.~\ref{fig:resnet-bilstm} outperforms other video classification models such as \ac{I3D} \cite{carreira2017quo} and \ac{R3D-18} \cite{tran2018closer}, which were fine-tuned with the same hyperparameters as our proposed model for the comparison. Our proposed model achieved an F1 score of 85.78\% on the test dataset. The model also performed worse when replacing the \ac{BiLSTM} layer with \ac{GRU} \cite{cho2014learning}, where ResNet-18 combined with \ac{GRU} achieved an F1 score of 19.34\%. It might be because \ac{BiLSTM} provides more contextual information. From Fig.~\ref{fig:bug-combination-metrics}, we can see the proposed model's performance on each bug combination from no bugs and individual bugs to up to bug combinations of 2 and 3 bug classes in a single moment in the video clips of the test dataset. Our model achieved an F1 score of 88.5\% as shown in Fig.~\ref{fig:bug-combination-metrics} when the video clip had a combination of the boundary hole, geometry corruption, and Z-clipping taking place simultaneously. It performed successfully when the clip had no bugs, with an F1 score of 95.6\%, and achieved an F1 score of 55.1\% on the clips with only the geometry corruption bug.

\section{Discussion}
From the results in Table~\ref{tab:model-comparison} and Fig.~\ref{fig:bug-combination-metrics}, we gained useful insights about the strengths and limitations of our proposed model architecture (ResNet-BiLSTM) on the proposed benchmark dataset. When comparing the use of \ac{BiLSTM} and \ac{GRU} for the temporal feature extraction in the models, we saw that the model performed better when using \ac{BiLSTM} with an F1 score gain of 76.44\%, while also outperforming \ac{I3D} \cite{carreira2017quo} and \ac{R3D-18} \cite{tran2018closer}. The likely reason is the ability of \ac{BiLSTM} to capture more contextual information while extracting temporal dependencies in both directions across frames, while other models might not leverage the temporal context as effectively.

Additionally, Fig.~\ref{fig:bug-combination-metrics} shows noticeable patterns where the model performs better with videos having bug combinations consisting of the boundary hole bug and moderately well with the Z-clipping bug. We also noticed that geometry corruption and Z-fighting bugs had the worst performance with the proposed model, even in a combination with other perceptual bugs when not including Z-clipping or boundary hole within the combination. This could be because geometry corruption and Z-fighting have subtler visual artifacts that are harder to distinguish than boundary hole or Z-clipping due to more pronounced visual anomalies.

Furthermore, we found that bug combinations with three classes of bugs performed better than with fewer bug combinations, which is counterintuitive. Although more bugs should be harder to detect, our model achieves better performance with more bugs, as more bugs in a video have more pronounced features that can make them easier to detect.

Moreover, the dataset has 65.55\% of video clips without any perceptual bugs to simulate real games with fewer instances of perceptual bugs. This could be the reason that the model performs well with no bugs present in the video. The bug occurrence frequency can be much lower depending on the video game. But we chose the mentioned distribution of bugs because of our limitation of needing a significant amount of generated buggy examples with limited computational resources.

On another note, our proposed dataset consists of only three games and five classes of perceptual bugs, which may limit generalization. Also, synthetic bugs injected into the gameplay session of our dataset may lack the nuance of real-world visual bugs, which are caused by broken underlying game logic, specific player interaction sequences, and environmental triggers in the game. Moreover, the perceptual bug combinations begin and end at the same time in the video. Varying durations of each class of buggy events would be beneficial to include for improving the dataset and testing our proposed model further.

\section{Conclusion and Future Work}
In this work, we introduced a new dataset for multi-label bug detection from perceptual bugs using gameplay videos. The dataset consisted of gameplay videos from three different games and included a combination of 1-3 types of bugs from 5 different classes of perceptual bugs in each frame in the gameplay videos. We compared different models on this dataset and found that our proposed model with ResNet and Bidirectional \ac{LSTM} achieved an F1 score of 85.78\%. Our model demonstrated that temporal context is necessary for capturing features needed to perform bug detection on videos.

Although our dataset includes thousands of video clips from three different games for multi-label bug detection, it is imperative to include more videos from different genres of games in future versions of this dataset. Future research should include an ablation study, real-time inference and temporal localization of these perceptual bug combinations for full-length gameplay. We hope that \ac{ABD} for multi-label perceptual bugs through our proposed approach can benefit both game developers and players by reducing quality assurance costs, accelerating development cycles, and improving game quality.



\bibliographystyle{IEEEtran}
\bibliography{references}

\end{document}